\documentclass{article}
\usepackage{spconf,amsmath,graphicx}
\usepackage{version}
\usepackage{caption}
\usepackage{color}
\usepackage{multirow}

\title{Superpixel-based Color Transfer}
\name{R{\'e}mi Giraud$^{1,2}$ \qquad Vinh-Thong Ta$^{1,3}$ \qquad Nicolas Papadakis$^{2}$
\thanks{This work has been carried out with financial support of the French 
State, managed by the French National Research Agency (ANR) in the  
frame of the GOTMI project (ANR-16-CE33-0010-01) and
the Investments for the future Program IdEx Bordeaux 
(ANR-10-IDEX-03-02) with the Cluster of excellence CPU.
}
}

\address{$^{1}$Univ. Bordeaux, LaBRI, CNRS, UMR 5800, F-33400 Talence, France.\\
    $^{2}$Univ. Bordeaux, IMB, CNRS, UMR 5251, F-33400 Talence, France.\\
     $^{3}$Bordeaux INP, LaBRI, UMR 5800, F-33405 Talence, France.
}

\begin{document}
\maketitle
\begin{abstract}

In this work, we propose a fast superpixel-based color transfer method (SCT)
between two images.
Superpixels enable to 
decrease the image dimension and to extract a reduced set of color candidates.
We propose to use a fast approximate nearest neighbor matching algorithm
in which we enforce the match diversity by limiting the selection of the same superpixels.
A fusion framework is designed to transfer the matched colors,
and we demonstrate the improvement obtained over exact matching results.
Finally,
we show that SCT is visually competitive compared to
state-of-the-art methods.

\end{abstract}

\section{Introduction}
\label{sec:intro}

Color transfer consists in modifying the color distribution
of a target image
using one or several reference source images.
The produced result must be
consistent with the target image structure, 
and computational time is an important issue to process large images or video sequences.

\noindent\textbf{Color transfer.} 
Initiated by \cite{Reinhard01}, many approaches have been proposed to 
transfer color statistics in different color spaces~\cite{Xiao:2006,pitie2007b,Papadakis_ip11,nguyen2014}. 
Optimal transportation tools have also been intensively studied to match and transfer the whole color distribution
~\cite{pitie2007a,Rabin_ip11,Frigo14}.
Nevertheless, as underlined in \cite{Reinhard11}, 
 since color distributions between images  may be very different, 
 the exact  transfer of the color palette 
 may produce visual outliers.
Relaxed optimal transport models, that do no exactly match color distributions,  
have then been proposed to tackle this issue \cite{Ferradans_siims13}, but they rely on time consuming algorithms.
  Moreover, when the process is only performed in the color space, incoherent colors may be transfered to neighboring pixels. 
  Artifacts such as JPEG compression blocks, enhanced noise or saturation then become visible~\cite{color_artifacts_suppression_14}, 
  unless considering object semantic information
  \cite{Frigo14}.
  In \cite{Tai2005},  an EM approach is used to estimate a Gaussian mixture model in color and pixel space,
  since the pixel location helps to preserve the image geometry.
However, a major limitation is the matching of clusters using a greedy approach based on nearest-neighbor criterion, 
with no control on the selected source colors.
In \cite{rabin2014non}, a relaxed optimal transport model is
applied to color transfer using superpixel lower-level representation.

\noindent\textbf{Superpixels.} These decomposition methods
reduce the image dimension by grouping pixels into homogeneous areas \cite{achanta2012,vandenbergh2012,giraud2016}.
They
have become widely used to reduce the 
computational burden of various image processing tasks such as 
multi-class object segmentation
\cite{tighe2010},  
object localization \cite{fulkerson2009} or
contour detection \cite{arbelaez2011}.
Generally,
the irregular geometry of the decompositions makes difficult their use into standard processes.
However, for color transfer application, superpixels become particularly interesting 
since they enable to describe consistent color areas, 
and matches can be found
regardless of the superpixel neighboring structure.

\newcommand{\iw}{0.155\textwidth}
\newcommand{\ih}{0.095\textwidth}
\begin{figure}[t!]
{\footnotesize
\begin{center}
\begin{tabular}{@{\hspace{0mm}}c@{\hspace{1mm}}c@{\hspace{1mm}}c@{\hspace{0mm}}}
 \includegraphics[width=\iw, height=\ih]{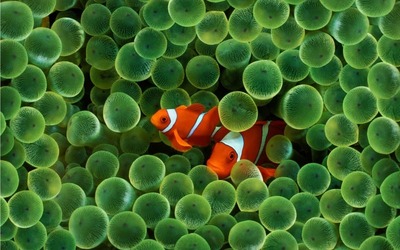} &  
 \includegraphics[width=\iw, height=\ih]{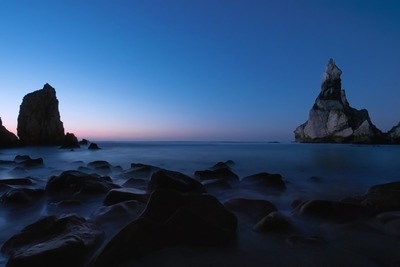}&
  \includegraphics[width=\iw, height=\ih]{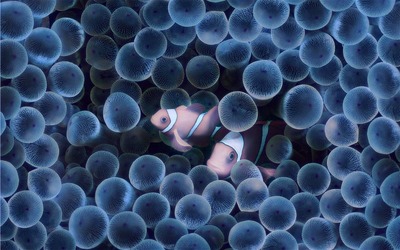}  \\
 Target image & Source image & SCT result\\
 \end{tabular} 
\end{center}
}
\caption{Example of color transfer with the proposed SCT method.
}\vspace{-0.4cm}
\label{fig:intro}
\end{figure}

Superpixel-based approaches such as \cite{rabin2014non,liu2016photo}
allow a better adaptation
to color histograms and image content, but still require
important computational cost.
Except for tracking applications \cite{wang2011},
fast superpixel matching methods have been little investigated.
For instance, in \cite{gould2012}, the PatchMatch (PM) algorithm \cite{barnes2009} 
that finds approximate nearest neighbor (ANN) patches
between images is adapted to graphs, and \cite{gould2014,giraud2017}
consider it in the superpixel context.

\noindent\textbf{Contributions.} 
In this work, we propose a fast superpixel-based color transfer method (SCT).
An example result of SCT is given in Figure \ref{fig:intro}.
To select the color candidates,
we use the fast and robust PM algorithm, 
that we adapt to handle superpixels \cite{giraud2017}. 
Contrary to \cite{Tai2005}, we propose a method to constrain the ANN search process
to limit the selection of the same superpixels in the source image.
Throughout the paper, we
demonstrate the significant improvement obtained with this constraint,
that enables to enforce the match diversity and to capture 
a larger color palette,
similarly to \cite{rabin2014non}. 
The selected colors are then transfered by a color fusion approach inspired from
the non-local means framework \cite{buades2005}.
Finally, we show that  SCT
produces accurate color transfer in low computational time thanks to the superpixel representation.

\section{Superpixel-based Color Transfer Method}

\subsection{ANN Superpixel Matching}

\textbf{PatchMatch algorithm.} 
The PatchMatch (PM) method \cite{barnes2009}
computes correspondences between pixel patches of
two images $A$ and $B$. 
It exploits the assumption that if 
patches are matched between $A$ and $B$, 
then their respective adjacent neighbors should also match well. 
Such propagation, associated with a random selection of patch candidates,
enables the algorithm to have a fast convergence towards good ANN.

PM is based on three steps.
The first one randomly assigns to each
patch of $A$, a corresponding patch in $B$.
An iterative refinement process %
is then performed  following a scan order (top left to bottom right)
to refine the correspondences with
the propagation and random search steps.
For a patch $A_i\in A$,
the aim is to find the match $B_{(i)}\in B$ that minimizes a
 distance $D(A_i,B_{(i)})$,
 for instance the sum of squares differences of color intensities.
During propagation, 
for each patch in $A$, the two recently processed  adjacent patches
are considered.
Their matches in $B$ are shifted to respect the relative positions in $A$, 
and the new candidates are tested for improvement.
Finally, the random search selects candidates around the current ANN in $B$
to escape from local minima. \smallskip

\noindent\textbf{Adaptation to superpixel matching.}
Several issues appear when considering the PM algorithm to the matching of superpixels \cite{giraud2017}.
Since superpixels decompose the image into irregular areas, 
there is no fixed adjacency relation between the elements.
First, a scan order must be defined to process the superpixels.
Then, during propagation, 
the decomposition geometry being also different in the image $B$, 
the shift of the neighbor cannot be directly performed.
A solution is to select the candidate with the most similar relative orientation 
computed with the superpixel barycenters.
Figure \ref{fig:prop} illustrates the selection of a candidate during propagation.
Such adaptation of PM provides 
a fast superpixel ANN matching algorithm
that produces an accurate selection of colors to transfer.

  \newcommand{\htabb}{0.125\textwidth}
    \newcommand{\wtabb}{0.47\textwidth}
  \begin{figure}[t!]
\begin{center}
 \begin{tabular}{@{\hspace{0mm}}c@{\hspace{0mm}}}
 \includegraphics[width=\wtabb,height=\htabb]{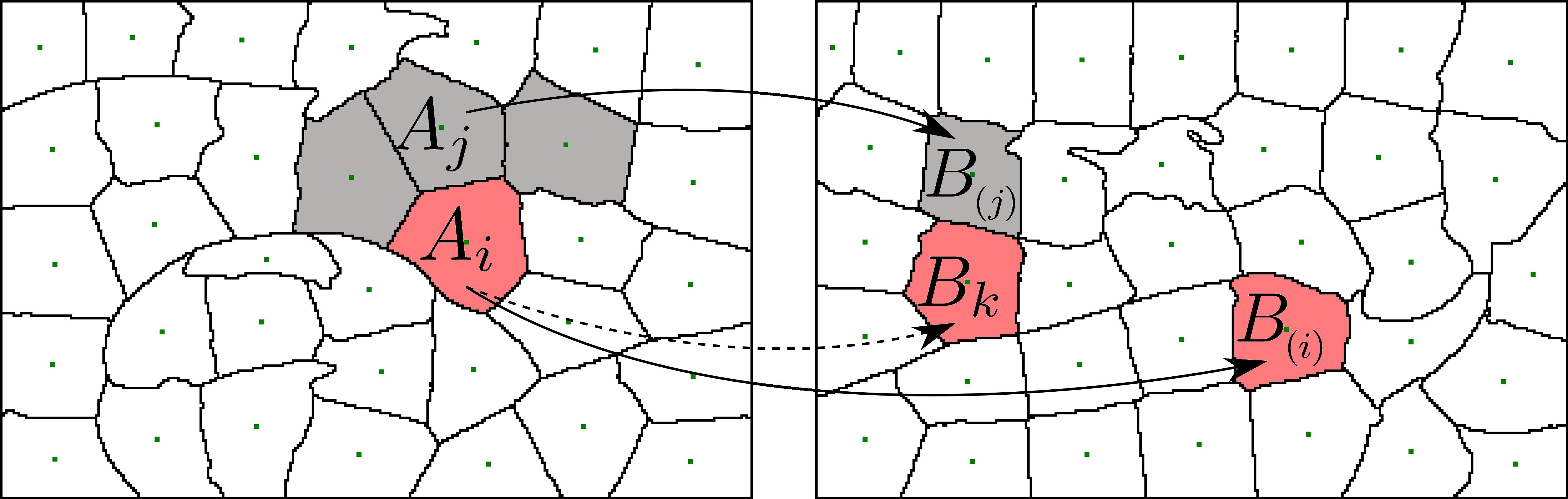}
  \end{tabular}
  \end{center}
  \caption{
    Illustration of propagation step.
    The superpixel $A_i$ (red) is currently matched to $B_{(i)}$.
    Its top-left adjacent neighbors $A_j$ (gray) are considered to provide new candidates.
    A neighbor  $A_j$ is matched to $B_{(j)}$, which leads to the candidate $B_k$,
    the neighbor of $B_{(j)}$ with the most similar relative position to the one between $A_i$ and $A_j$.
  } \vspace{-0.4cm}
  \label{fig:prop}
  \end{figure}

\subsection{Constraint on Match Diversity}

\begin{figure}[t!]
\begin{center}
{\footnotesize
\begin{tabular}{@{\hspace{0mm}}c@{\hspace{1mm}}}
\begin{tabular}{@{\hspace{0mm}}c@{\hspace{1mm}}c@{\hspace{1mm}}}
\includegraphics[width=0.11\textwidth]{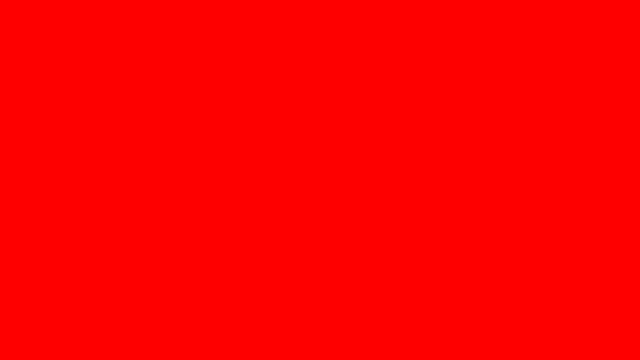}&
\includegraphics[width=0.11\textwidth, height=0.08\textwidth]{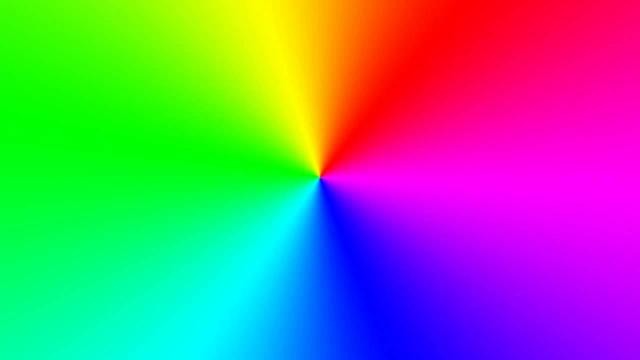}\\
 Target image & Source image \\
 \end{tabular} 
 \begin{tabular}{@{\hspace{0mm}}c@{\hspace{0mm}}}  \vspace{-0.35cm}
 \\ 
 ($\epsilon=\infty$) \\
 \includegraphics[width=0.11\textwidth]{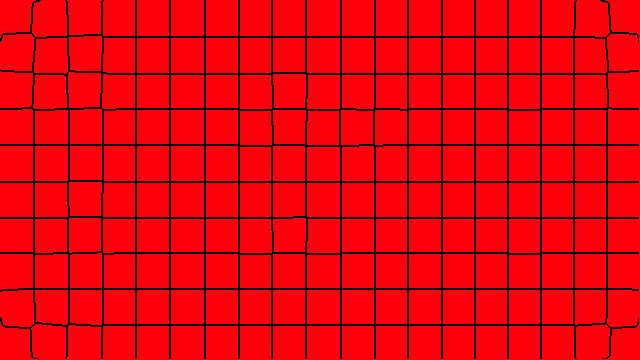}\\
 Matched colors\\
 \end{tabular}
  \begin{tabular}{@{\hspace{0mm}}c@{\hspace{0mm}}}   \vspace{-0.35cm}
  \\
   ($\epsilon=1$) \\
   \includegraphics[width=0.11\textwidth]{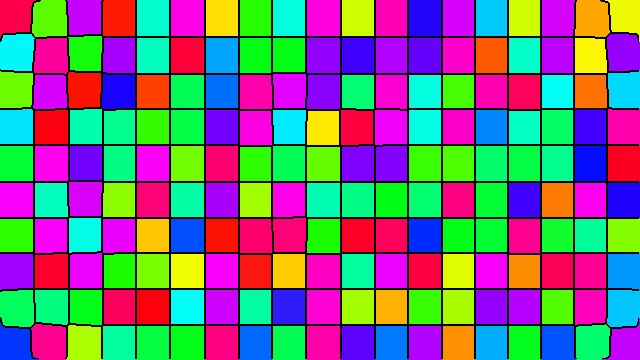}\\
   Matched colors\\
  \end{tabular}
  \end{tabular}
  }
  \end{center}
  \caption{Illustration of matching without ($\epsilon=\infty$) and with ($\epsilon=1$)
  constraint on the number of source superpixel selection. Without constraint, all target
  superpixels match a red one in the source image.} \vspace{-0.45cm}
  \label{fig:mire_ex} 
  \end{figure}

In practice,
the ANN search
may converge towards exact matching, almost providing the nearest neighbors.
The aim of  color transfer is to globally capture the source color palette.
If the source image contains one or several 
elements that match well the color set of the target image $A$,
the ANN search may lead to the same match in $B$.
The color transfer would thus provide a result very close to $A$.
Figure \ref{fig:mire_ex} illustrates this issue.
Since the source image also contains red colors,
all superpixels of the target image 
find a close red match in the source space,
leading to no color transfer.

To  enforce the match diversity and capture a larger color palette of the source image,
we propose
to constrain the ANN search 
and to restrict the number of associations to the same element.
To do so, we set
a parameter $\epsilon$ that defines the maximum number of selection
of the same superpixel.
Such constraint requires the number of source elements $|B|$ to be such that
$|A|\leq\epsilon|B|$.
In Figure \ref{fig:mire_ex}, with $\epsilon=1$,
the target superpixels
now capture the global palette of the source image. 
First, we make sure that the initialization step respects this constraint when
randomly assigning the correspondences.
Then, during the following iterative process,
a superpixel $A_i$ can be assigned to a superpixel $B_k$, 
only if less than $\epsilon$ superpixels in $A$ are already assigned to $B_k$.
If $B_k$ is already matched by $\epsilon$ elements in $A$,
one superpixel $A_j$ assigned to $B_k$ must be 
sent to another superpixel in $B$ to 
allow $A_i$ to match $B_k$.
We propose to compute the cost of sending a superpixel $A_j$, currently matched with $B_k$,
towards $B_{(i)}$, the current correspondence of $A_i$, thus making a switch between the matches,
and ensuring the respect of the constraint set by $\epsilon$.
For all superpixels $A_j$ matched to $B_{(j)}=B_k$,
the switching cost is considered
as follows: \vspace{-0.15cm}
\begin{align} 
C(A_i,A_j)   = ( &D(A_i,B_{(j)}) -  D(A_i,B_{(i)})) \nonumber  \\
                           &+  (D(A_j,B_{(i)}) - D(A_j,B_{(j)})) .
\end{align}
\noindent
If a superpixel $A_j$ reduces the global matching distance, \emph{i.e.},
if the cost $C<0$,
we proceed to the following assignments:
$\underset{A_j}{\text{argmin}}(C(A_i,A_j)) \rightarrow B_{(i)}$ and
$A_i \rightarrow B_{(j)}=B_k$. 

\noindent\textbf{Comparison to optimal assignment.}
The proposed ANN algorithm with
$\epsilon=1$
approximates the optimal assignment problem,
addressed with the
Hungarian or Munkres algorithms \cite{munkres1957}.
Given two sets of elements $\{A_i\}_{i\in\{1,\dots,|A|\}}$
and $\{B_j\}_{j\in\{1,\dots,|B|\}}$ with $|A|\leq|B|$,
the aim is to find to each $A_i$, an assignment $B_{(i)}$ that
can only be selected once, and to 
minimize the global distance between the matched elements.

In Figure \ref{fig:hung}, 
we consider two close images of $1920{\times}800$ pixels \cite{korman2011}.
We show the target image reconstruction $\widetilde{A}$ from the  
matched superpixels average colors  and compare total matching distance and computational time
between our approach, optimal resolution \cite{munkres1957}, and
to random superpixel assignments, \emph{i.e.}, 
the initialization step of our algorithm.
Our fast superpixel ANN method provides close results to  the optimal resolution
while being order of magnitude faster.

    \newcommand{\hhtabsf}{0.095\textwidth}
    \newcommand{\wwtabsf}{0.21\textwidth}
        \newcommand{\hhtabsff}{0.125\textwidth}
    \newcommand{\wwtabsff}{0.23\textwidth}
  \begin{figure}[t!]
\begin{center}
{\scriptsize
 \begin{tabular}{@{\hspace{0mm}}c@{\hspace{2mm}}c@{\hspace{0mm}}}
 \includegraphics[width=\wwtabsf,height=\hhtabsf]{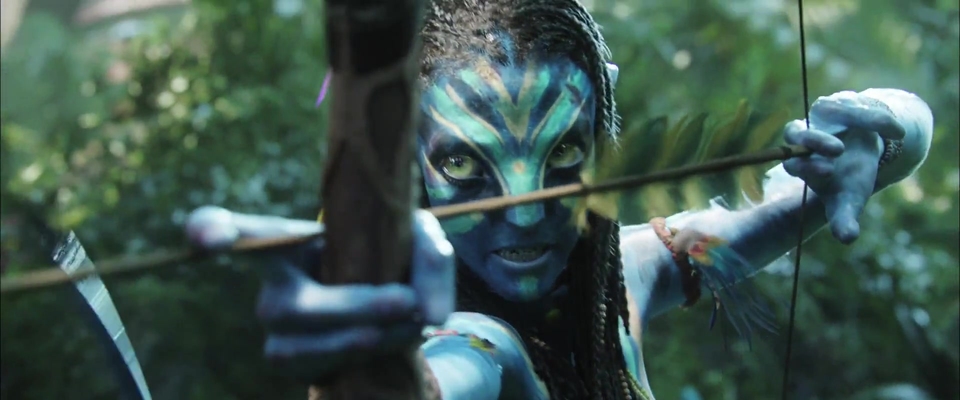}&
 \includegraphics[width=\wwtabsf,height=\hhtabsf]{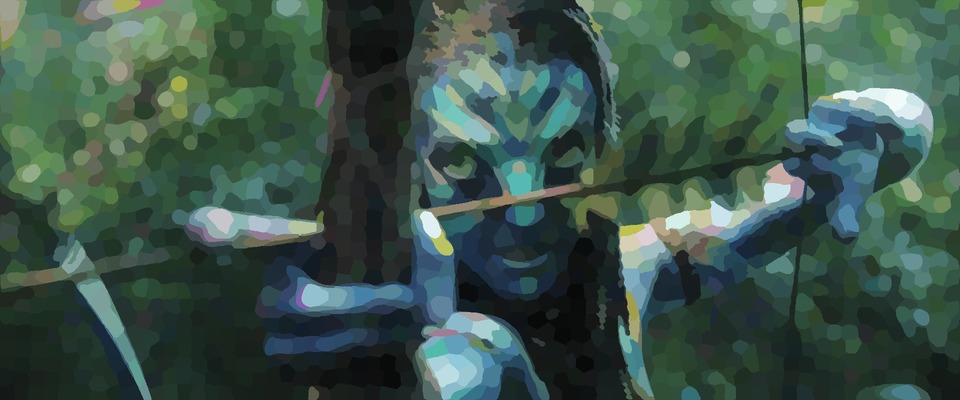}\\
     Target image $A$& $\widetilde{A}$ from \cite{munkres1957}\\
         \includegraphics[width=\wwtabsf,height=\hhtabsf]{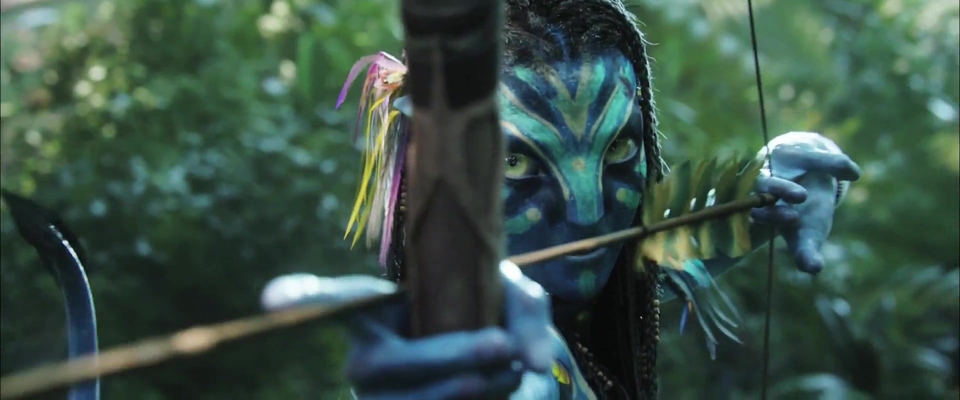}&
    \includegraphics[width=\wwtabsf,height=\hhtabsf]{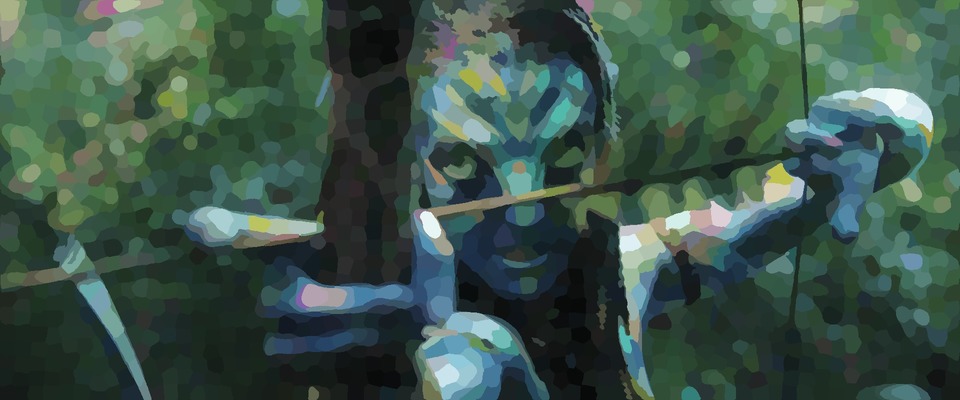}\\
     Source image $B$& $\widetilde{A}$ from our superpixel ANN matching\\
  \end{tabular}
   \begin{tabular}{@{\hspace{0mm}}c@{\hspace{1mm}}c@{\hspace{0mm}}}
 \includegraphics[width=\wwtabsff,height=\hhtabsff]{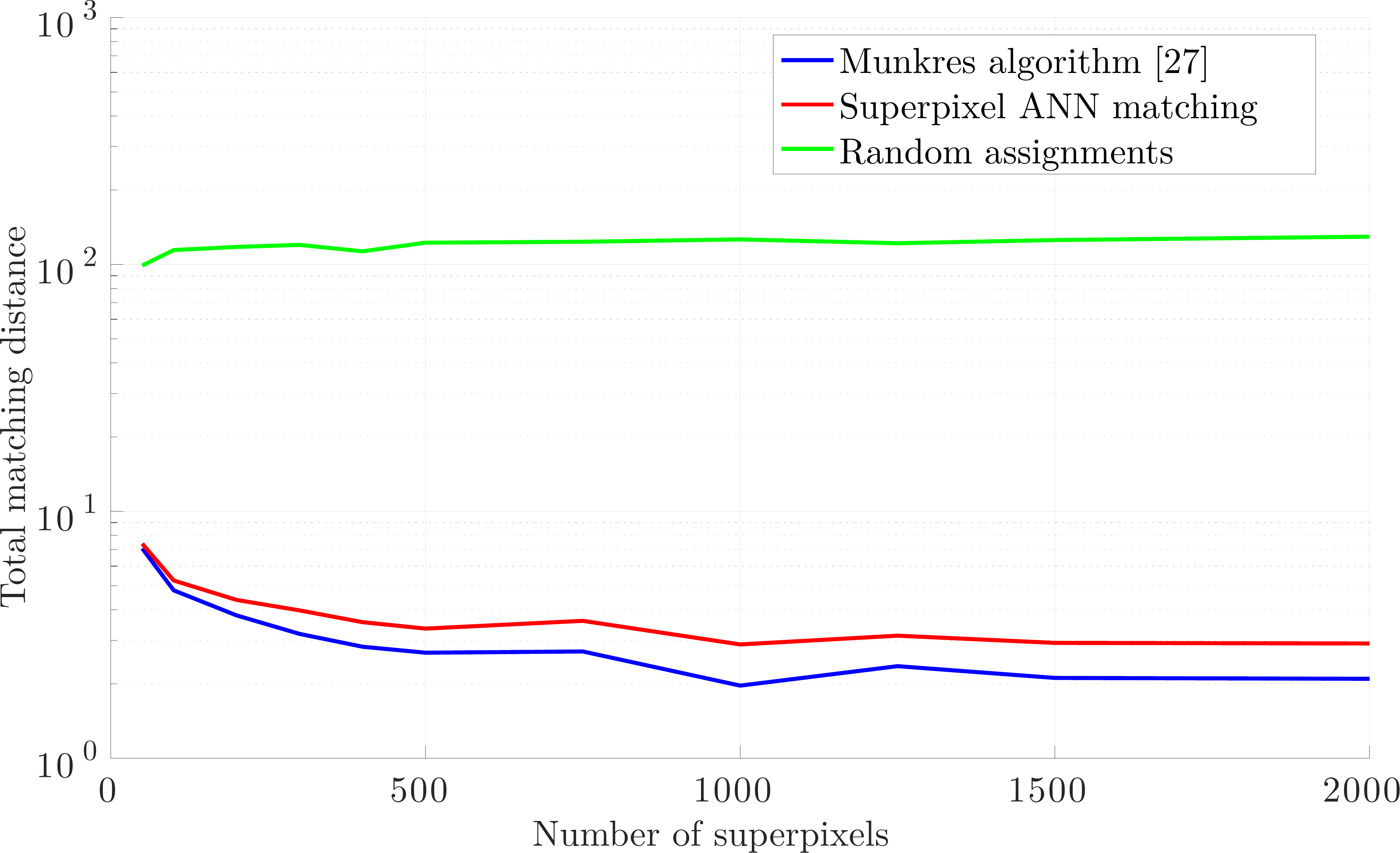}&
    \includegraphics[width=\wwtabsff,height=\hhtabsff]{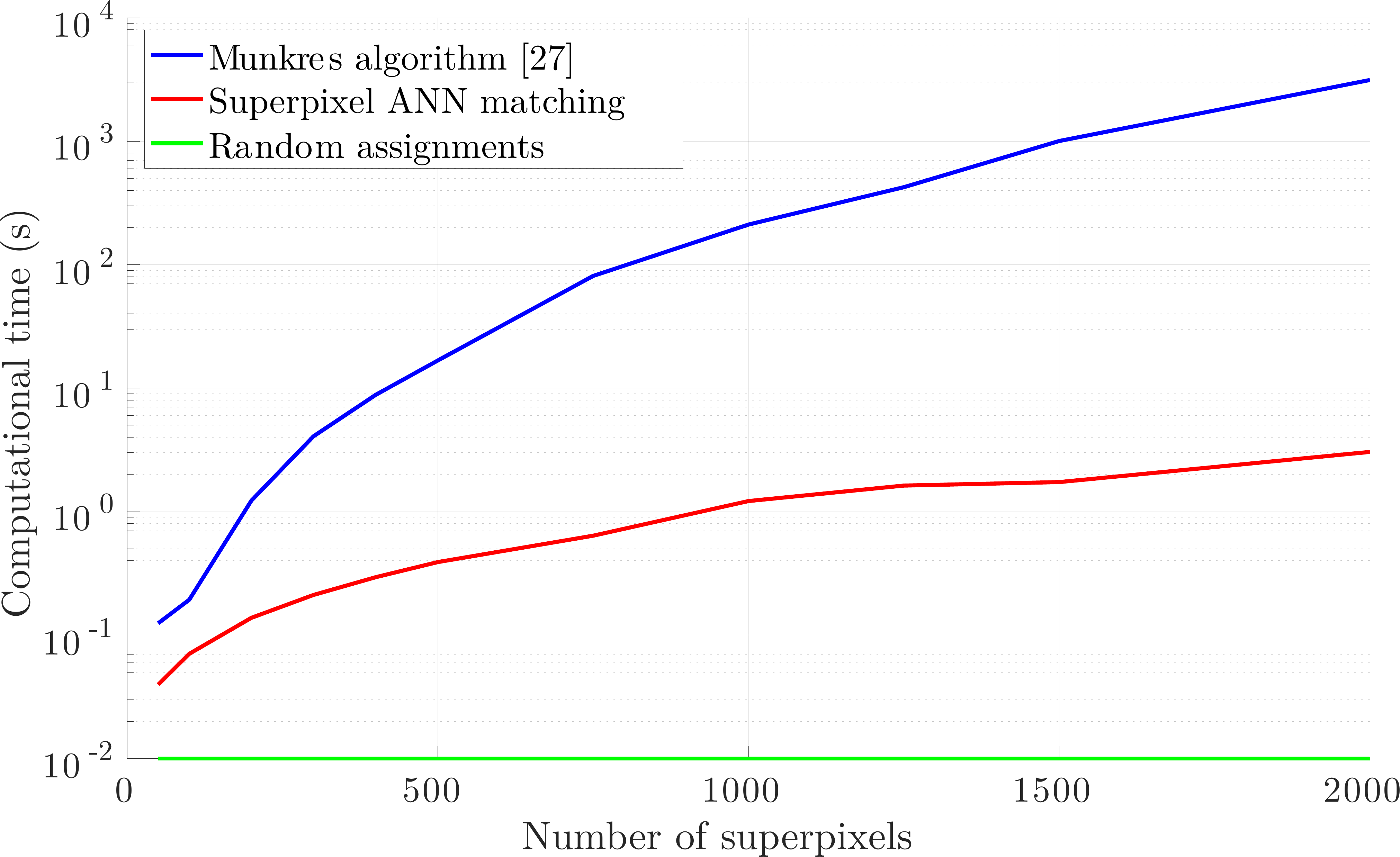}\\
  \end{tabular}
  }
  \end{center}
  \caption{
  Comparison of optimal assignment using Munkres algorithm \cite{munkres1957} and
  our constrained superpixel ANN matching ($\epsilon=1$).
  Target image reconstruction from the
  matched superpixels  average colors  ($K\approx2000$) and
  performances for several superpixel scales.
  } \vspace{-0.4cm}
  \label{fig:hung}
  \end{figure}

\subsection{Color Fusion Framework}

Our matching framework provides an ANN in $B$ to each superpixel of $A$.
The aim is then to transfer the color of the matches to compute
the color transfer image $A_t$ while preserving the structure of the target $A$.
Since the matched superpixels are very likely to have different shapes, 
there are no direct pixel associations between elements, 
and source colors cannot be directly transfered to $A$ at the pixel scale.
The average colors of superpixels in $B$ can be transfered
to the ones in $A$ but it
would give a piece-wise color transfer result.

We propose to consider the average colors from the matched superpixels
in a non-local means fusion framework \cite{buades2005}.
Hence, all matched colors can contribute to the color computation of each pixel in $A_t$.
Such approach enables to increase the number of color candidates
and leads to new potential ones that adapt well to the target image content.
A superpixel $A_i$ is described by the set of positions $X_i$ and colors $C_i$ of the 
contained pixels $p$, such that $A_i = [X_i,C_i] = [(x_i/N_x,y_i/N_y),(r_i,g_i,b_i)/255]$,
with $N_x{\times}N_y$ the size of image $A$.
To compute the new color $A_t(p)$ of a pixel $p$,
the weighted fusion of the matched colors is performed 
based on color and spatial similarity:  \vspace{-0.1cm}
\begin{equation}
A_t(p) = \frac{\sum_{j} \omega(p,A_j)\bar{C}_{B_{(j)}}}{\sum_{j} \omega(p,A_j)}   ,  \label{fusion_sp}
\end{equation}
with 
$\bar{C}_{B_{(j)}}$, the average color of the match of $A_j$ in $B$, 
and
$\omega(p,A_j)$ the weight that depends on the 
distance between the considered pixel $p\in A_i$ and the superpixel $A_j\in A$.
This weight is computed similarly to a Mahalanobis distance: \vspace{-0.1cm}
\begin{equation}
 \omega(p,A_j) = \exp{\left( -\left((p-\bar{A_j})^T Q_i^{-1} (p-\bar{A_j}) - \sigma(p) \right)\right) ,}
\end{equation}
where $\sigma(p)$ sets the exponential dynamic and is set such
that
 $\sigma(p) = \underset{j}{\min}{\left( (p-\bar{A_j})^T Q_i^{-1} (p-\bar{A_j}) \right) } $, and
$Q_i$ includes spatial and colorimetric
covariances of the pixels in
$A_i$: \vspace{-0.1cm}
\begin{equation}
 Q_i=Q(A_i)=\begin{pmatrix}
   \delta_s^2Cov(X_i) & 0\\
   0 & \delta_c^2Cov(C_i)\\
\end{pmatrix}. 
\label{Q}
\end{equation}

    \newcommand{\hhtabs}{0.10\textwidth}
    \newcommand{\wwtabs}{0.155\textwidth}
  \begin{figure}[t!]
\begin{center}
{\footnotesize
 \begin{tabular}{@{\hspace{0mm}}c@{\hspace{2mm}}c@{\hspace{1mm}}c@{\hspace{0mm}}}
 \includegraphics[width=\wwtabs,height=\hhtabs]{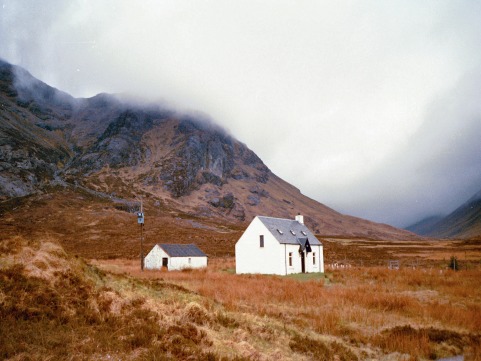}&
    \includegraphics[width=\wwtabs,height=\hhtabs]{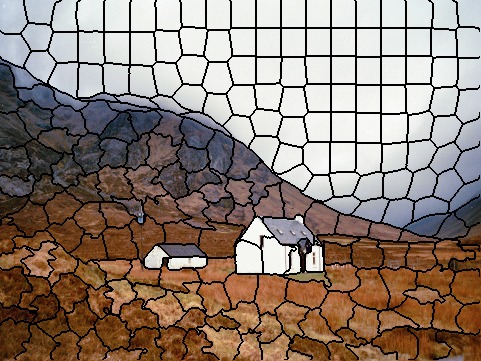}&
     \includegraphics[width=\wwtabs,height=\hhtabs]{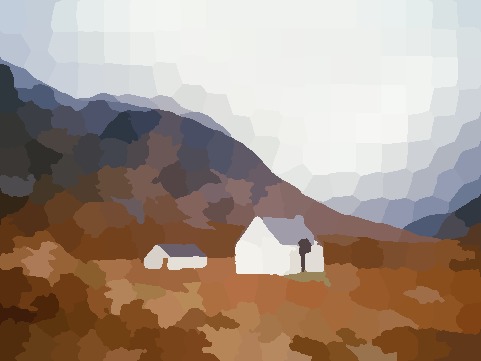}\\
 \includegraphics[width=\wwtabs,height=\hhtabs]{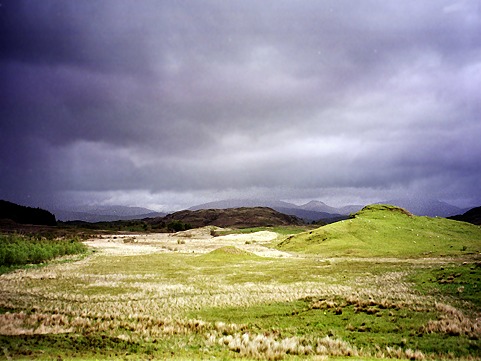}&
    \includegraphics[width=\wwtabs,height=\hhtabs]{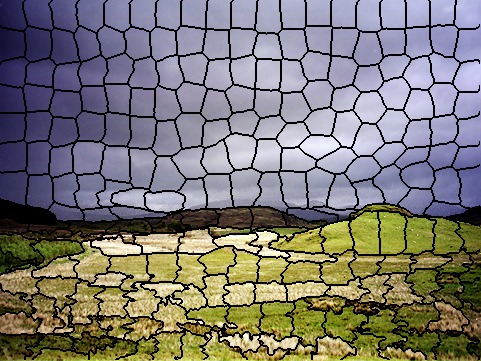}&
     \includegraphics[width=\wwtabs,height=\hhtabs]{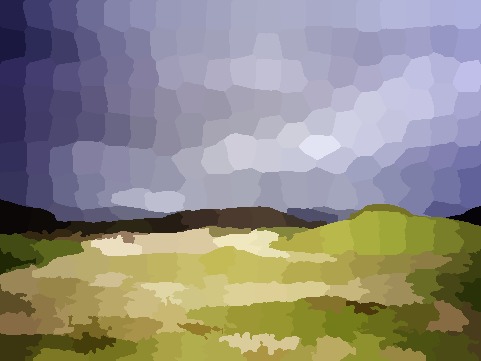}\\
    Target/Source images&Superpixels&Average colors\\
{\includegraphics[width=\wwtabs,height=\hhtabs]{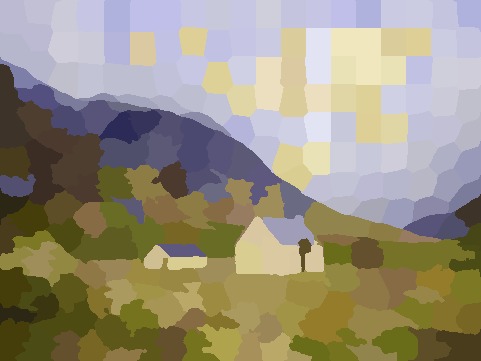}}&
    {\includegraphics[width=\wwtabs,height=\hhtabs]{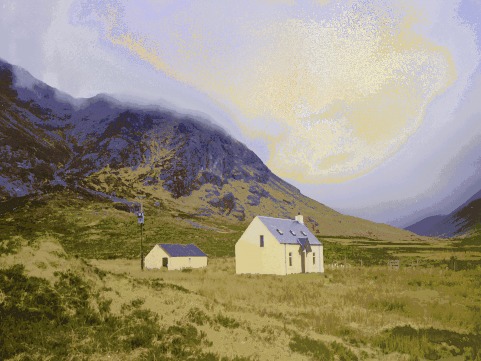}}&
 { \includegraphics[width=\wwtabs,height=\hhtabs]{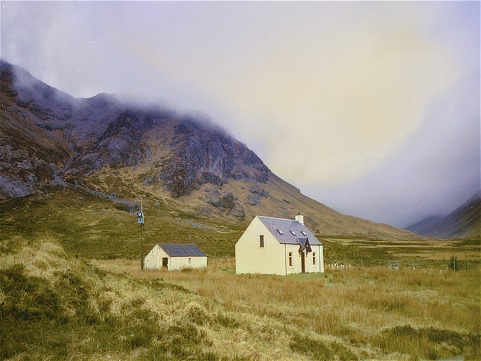}}\\ 
 Matched colors& Color fusion& Color fusion + \cite{pitie2005r}\\
  \end{tabular}
  }
  \end{center}
  \caption{
  Results of each SCT step. See text for more details.
  } \vspace{-0.1cm}
  \label{fig:method}
  \end{figure}

The SCT steps
are illustrated in Figure \ref{fig:method}.
We show the decomposition of
images into superpixels with average colors,
the matched source colors, and results of color fusion and
post-processing with a color regrain \cite{pitie2005r}.

\section{Color Transfer Results}

\subsection{\label{param_settings}Parameter Settings}

SCT is implemented with MATLAB using C-MEX code.
Superpixel decompositions are computed using \cite{giraud2016} such that
each superpixel approximately contains $500$ pixels.
The superpixel matching is performed on 
normalized cumulative color histogram features
and the number of ANN search 
iterations is set to $20$. 
The covariance parameters in Eq. \eqref{Q} 
are set such that $\delta_s$=$100\delta_c$ and $\delta_c$=$0.1$
in order to favor spatial consistency.
Finally, unless mentioned, $\epsilon$ is set to $3$ and 
the results are slightly refined 
using a color regrain \cite{pitie2005r}.

Our method produces results in very low computational time
due to the use of superpixels, \emph{i.e.}, 
less than $1$s for images of $480{\times}360$ pixels. 
Decompositions are computed with \cite{giraud2016} in less than $0.4$s,
matching is performed in approximately $0.1$s
and color fusion takes $0.25$s to provide the color transfer.

     \newcommand{\hhtabtt}{0.11\textwidth}
   \newcommand{\hhtabt}{0.11\textwidth}
    \newcommand{\wwtabt}{0.15\textwidth}
  \newcommand{\hhtabb}{0.11\textwidth}
    \newcommand{\wwtabb}{0.15\textwidth}
  \begin{figure*}[t!]
\begin{center}
{\footnotesize
 \begin{tabular}{@{\hspace{0mm}}c@{\hspace{1mm}}c@{\hspace{1mm}}c@{\hspace{1mm}}c@{\hspace{0mm}}}  
\includegraphics[width=0.16\textwidth,height=\hhtabb]{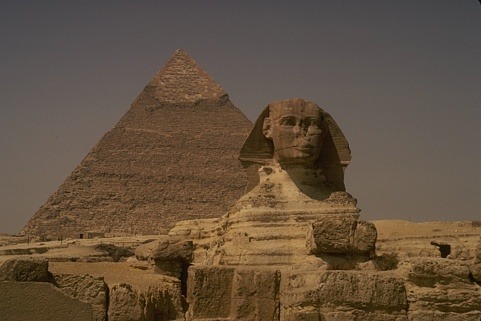}&
   \includegraphics[width=0.16\textwidth,height=\hhtabb]{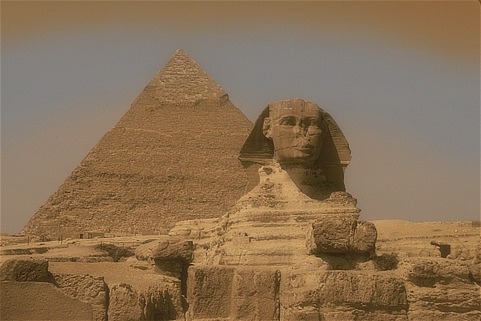}&
   \includegraphics[width=0.16\textwidth,height=\hhtabb]{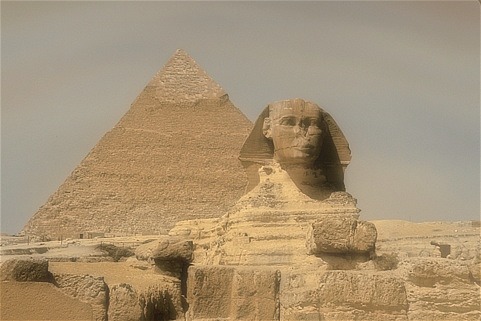}\\
   \includegraphics[width=0.16\textwidth,height=\hhtabt]{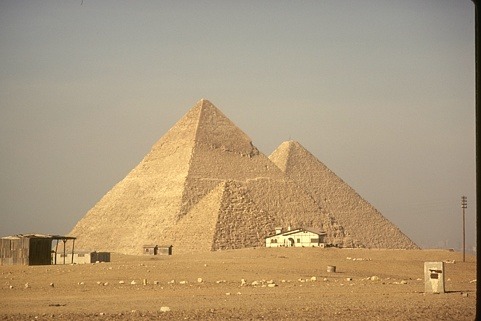}&
     \includegraphics[width=0.16\textwidth,height=\hhtabt]{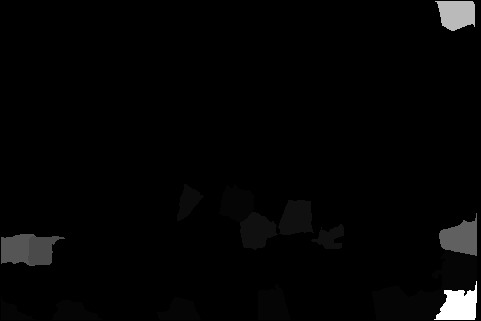}&
       \includegraphics[width=0.16\textwidth,height=\hhtabt]{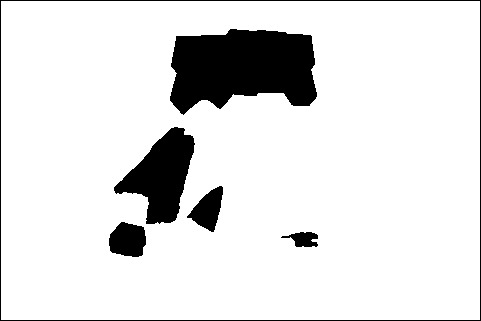}\\
 Target/Source images & SCT result $\epsilon=\infty$& SCT result $\epsilon=1$ \\
  \end{tabular}
  } 
  {\footnotesize
 \begin{tabular}{@{\hspace{0mm}}c@{\hspace{1mm}}c@{\hspace{1mm}}c@{\hspace{1mm}}c@{\hspace{0mm}}} 
   \includegraphics[width=\wwtabb,height=\hhtabb]{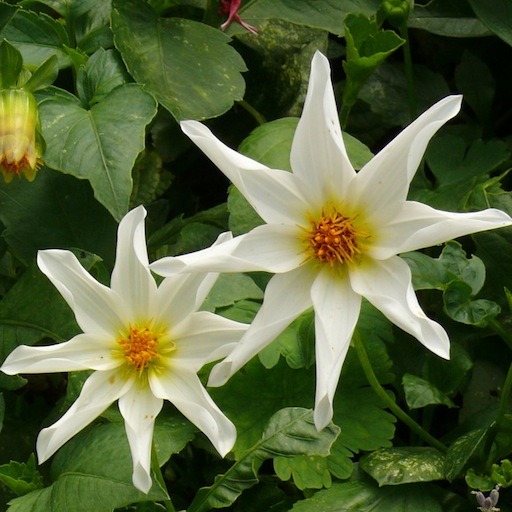}&  
   \includegraphics[width=\wwtabb,height=\hhtabb]{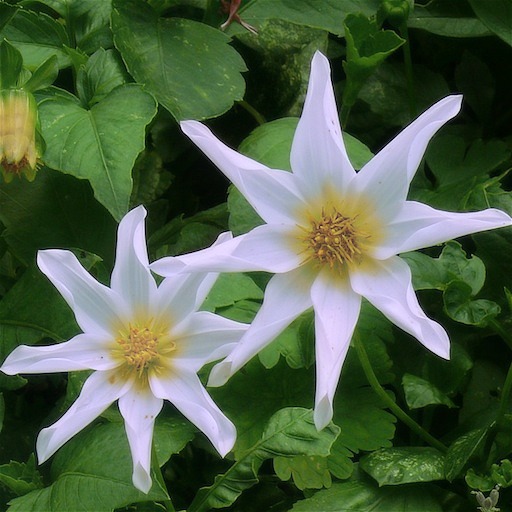}&
      \includegraphics[width=\wwtabb,height=\hhtabb]{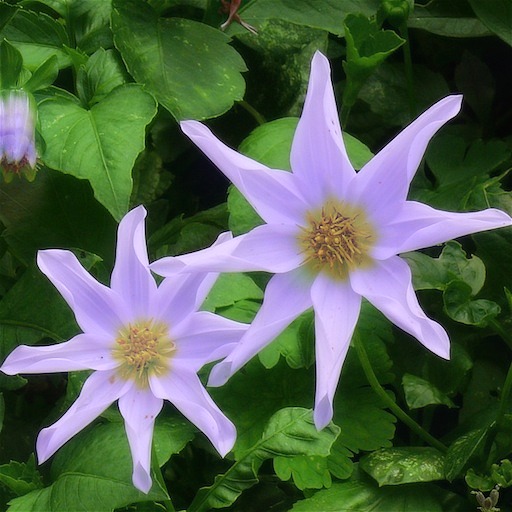}\\
 \includegraphics[height=\hhtabtt]{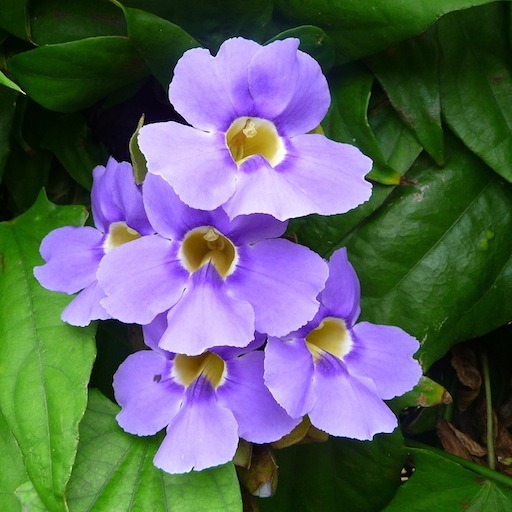}&
  \includegraphics[height=\hhtabtt]{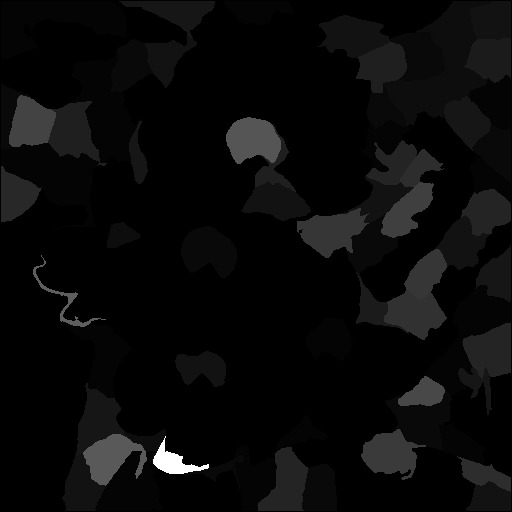}&
  \includegraphics[height=\hhtabtt]{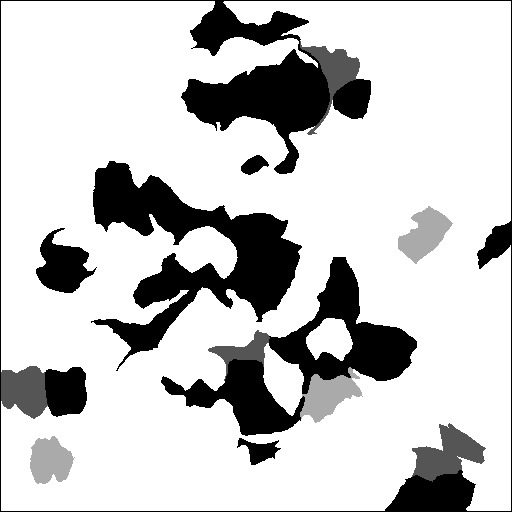}\\
 Target/Source images & SCT result $\epsilon=\infty$ & SCT result $\epsilon=3$ \\
  \end{tabular}
  }
  \end{center}
  \caption{
  Examples of color transfer results are shown for different $\epsilon$ values and
  compared to the results obtained with no constraint ($\epsilon=\infty$).
  The maps (bottom row) indicate the number of selection of the source superpixels 
  (black is zero and white is the highest number of selection).
  } \vspace{-0.15cm}
  \label{fig:influ_divers}
  \end{figure*}

   \newcommand{\wwtabbbi}{0.12\textwidth}
    \newcommand{\wwtabbb}{0.1675\textwidth}
        \newcommand{\hjh}{0.11\textwidth}
        \newcommand{\hjhh}{0.11\textwidth}
    \begin{figure*}[ht!]
\begin{center}
{\footnotesize
 \begin{tabular}{@{\hspace{0mm}}c@{\hspace{1mm}}c@{\hspace{1mm}}c@{\hspace{1mm}}c@{\hspace{1mm}}c@{\hspace{1mm}}c@{\hspace{1mm}}c@{\hspace{0mm}}}
  \includegraphics[width=\wwtabbb,height=\hjh]{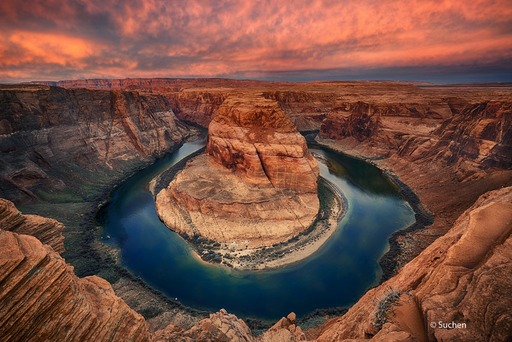}&
  \includegraphics[width=\wwtabbbi]{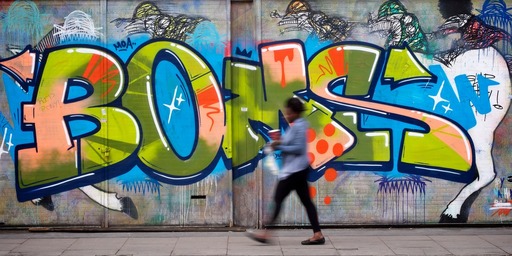}&
  \includegraphics[width=\wwtabbb,height=\hjh]{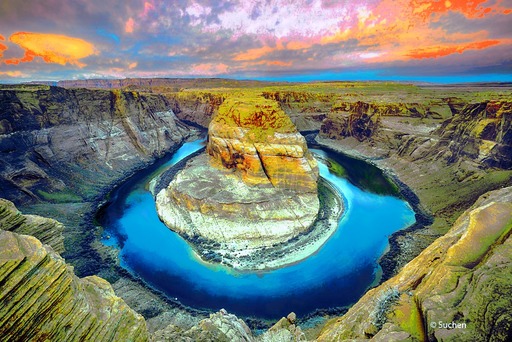}&
  \includegraphics[width=\wwtabbb,height=\hjh]{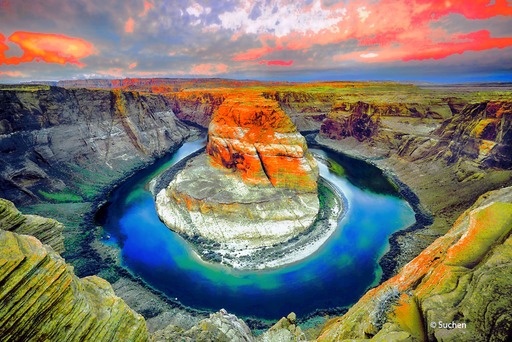}&
  \includegraphics[width=\wwtabbb,height=\hjh]{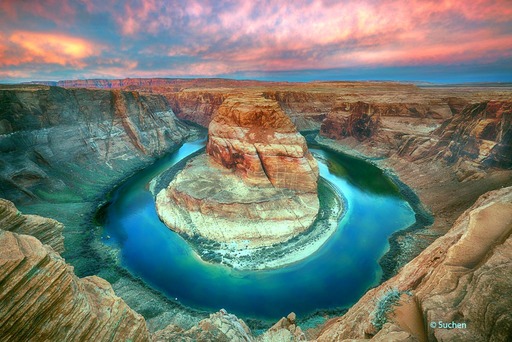}&
  \includegraphics[width=\wwtabbb,height=\hjh]{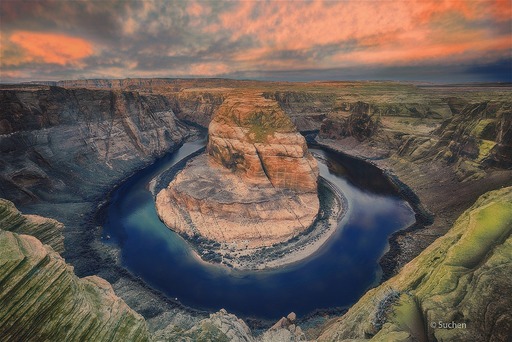}\\
   \includegraphics[width=\wwtabbb]{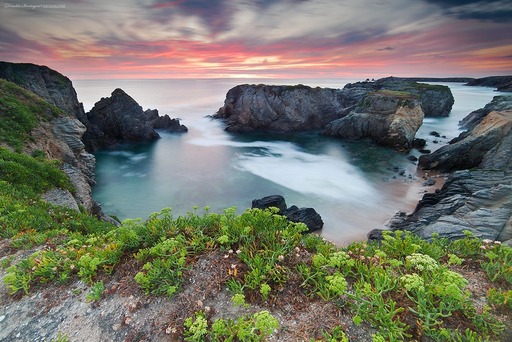}&
   \includegraphics[width=\wwtabbbi]{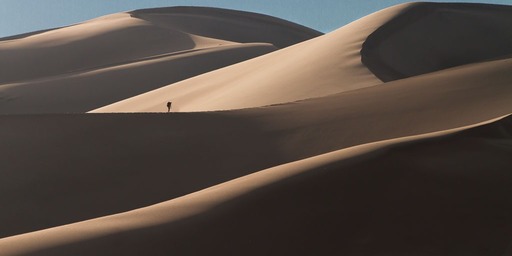}&
  \includegraphics[width=\wwtabbb]{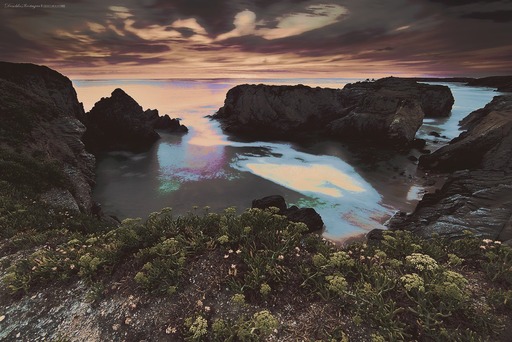}&
  \includegraphics[width=\wwtabbb]{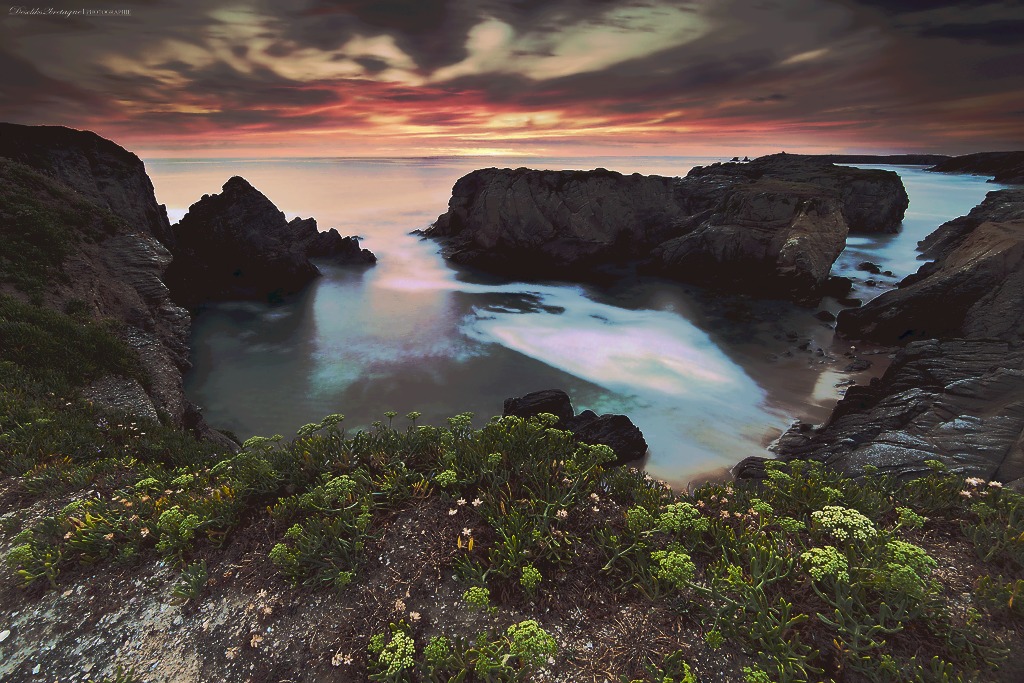}&
  \includegraphics[width=\wwtabbb]{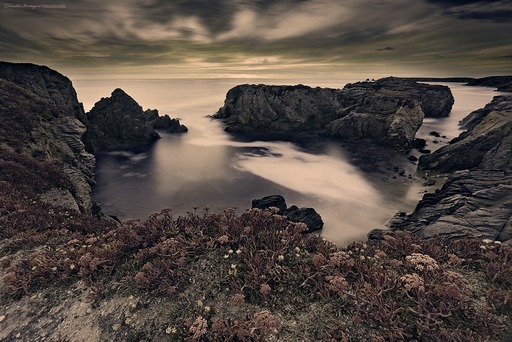}&
  \includegraphics[width=\wwtabbb]{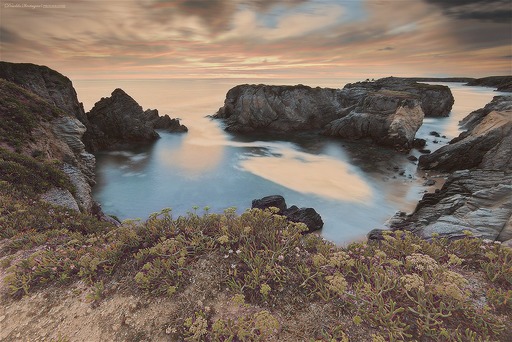}\\
  \includegraphics[width=\wwtabbb,height=\hjhh]{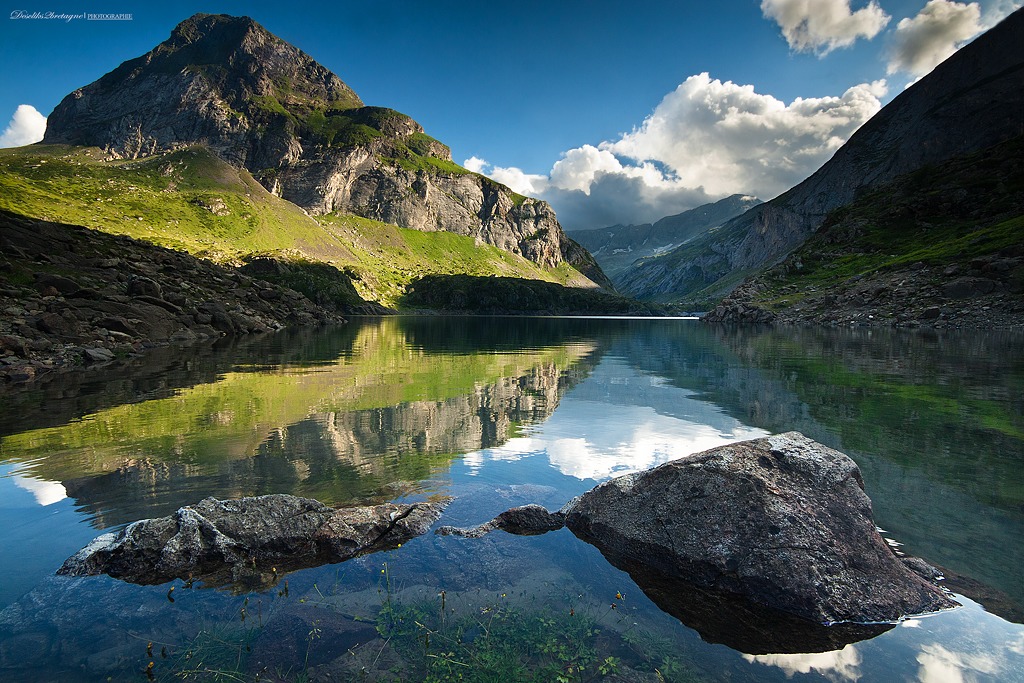}&
  \includegraphics[width=\wwtabbbi]{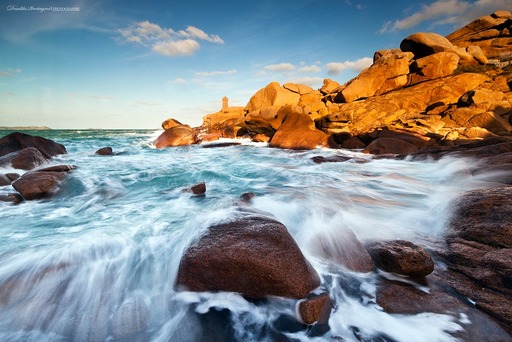}&
  \includegraphics[width=\wwtabbb,height=\hjhh]{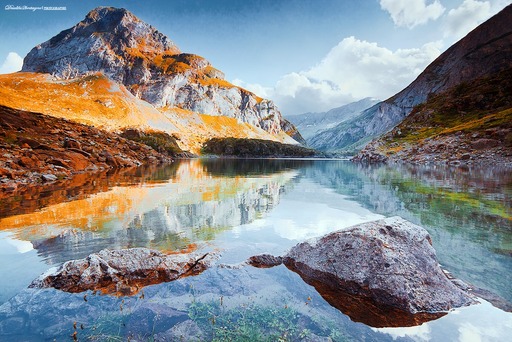}&
  \includegraphics[width=\wwtabbb,height=\hjhh]{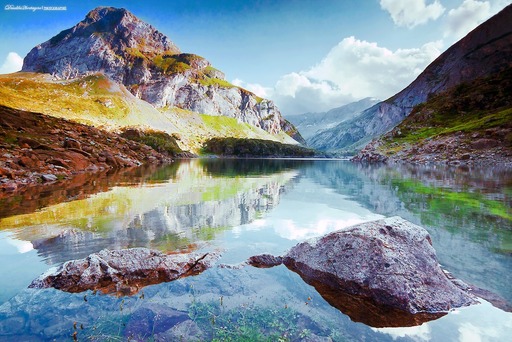}&
  \includegraphics[width=\wwtabbb,height=\hjhh]{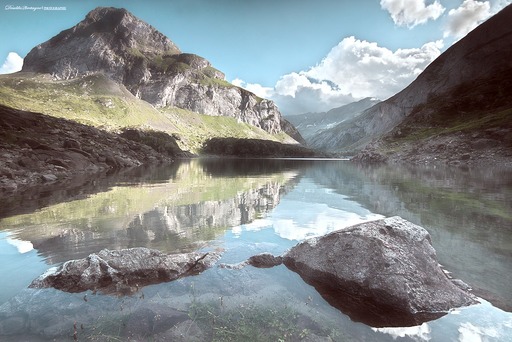}&
  \includegraphics[width=\wwtabbb,height=\hjhh]{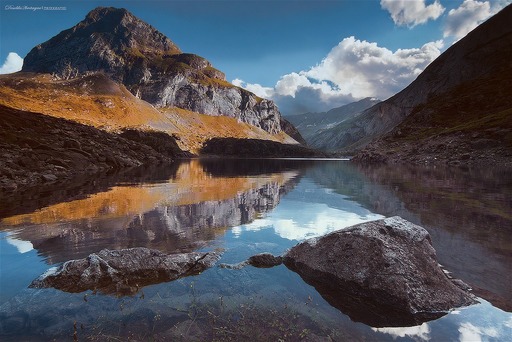}\\
  Target image&Source image&   \cite{pitie2007a}  & \cite{Papadakis_ip11} & \cite{nguyen2014} & SCT\\
  \end{tabular}
  }
  \end{center}
  \caption{
  Visual comparison to  \cite{pitie2007a}, \cite{Papadakis_ip11} and \cite{nguyen2014}.
  SCT provides more visually satisfying or equivalent results to the compared methods.
  } \vspace{-0.2cm}
  \label{fig:comp}
  \end{figure*}

\subsection{\label{influ_params}Influence of Match Diversity}

Additionally to Figure \ref{fig:mire_ex}, 
we illustrate in Figure \ref{fig:influ_divers} the influence of the $\epsilon$ parameter
that limits the
source superpixel selection.
It appears that
even fast ANN search may lead to the same best match,
as shown with the maps
corresponding to the selection of source superpixels.
For instance, 
most target superpixels of the white flower 
match the only white superpixel in the source image, leading to
almost no color transfer.
With the proposed method, we select accurate matches while capturing the global color palette 
of the source image.

\subsection{{\label{soa}}Comparison with State-of-the-Art Methods}

In this section, we compare the results of SCT to 
various methods based on
optimal transport \cite{pitie2007a},
histogram transfer with a variational model \cite{Papadakis_ip11} 
and 3D color gamut mapping \cite{nguyen2014}.
Figure \ref{fig:comp} illustrates color transfer examples for all methods.

SCT produces more visually satisfying
results than the ones of the compared methods.
The colors are relevantly transfered to the target image with
respect to the initial grain and exposure.
For instance, on the first image (top row),
\cite{pitie2007a}, \cite{Papadakis_ip11} and \cite{nguyen2014}
produce color transfers that strongly modify the illumination of the target image.
All compared methods except SCT fail at transferring the blue color from the desert sky into the sea (middle row),
and the orange color of the stones
to the grass of the moutain
(bottom row).
Contrary to the compared methods, we consider a selection of the source colors 
and our fusion model enables to adapt to the target image, preserving its 
structure and initial exposure.
Finally, 
while SCT results are computed in less than $1$s, while
other models such as \cite{Papadakis_ip11} may require 
prohibitive computational times, up to $120$s.

\section{Conclusion}

In this work, we propose a novel superpixel-based method for color transfer.
Our algorithm is based on a fast ANN search and fusion of source colors in a non-local means framework.
We introduce a method to constraint the neighbors diversity in the matching process,
to get a large color palette of source superpixels.
The colors are globally transfered to the target image with
respect to the initial grain and exposure, producing
visually consistent results.
Finally, the use of superpixels within our framework
enables to produce color transfer
in very limited computational time.
 Future works will
 focus on the adaptation
 to video color transfer using supervoxels.

   \newpage
\bibliographystyle{IEEEbib}
\bibliography{ICIP,refs}

\end{document}